%% file: paper.tex
\documentclass[10pt,twocolumn,letterpaper]{article}

\usepackage{iccv}
\usepackage{times}
\usepackage{epsfig}
\usepackage{graphicx}
\usepackage{amsmath}
\usepackage{amssymb}
\usepackage{caption}
\usepackage{multirow}
\usepackage{xcolor}
\usepackage{pifont}%
\usepackage{booktabs}
\usepackage{balance}
\usepackage[numbers,sort,compress]{natbib}

\usepackage{array}

\usepackage{diagbox}

\usepackage{tabularx}

\usepackage[pagebackref=true,breaklinks=true,letterpaper=true,colorlinks,bookmarks=false]{hyperref}

\usepackage[noabbrev,capitalise]{cleveref}
\usepackage{subcaption}

\newcommand{\cmark}{\ding{51}}%
\newcommand{\xmark}{\ding{55}}%
\newcommand{\qheading}[1]{\noindent\textbf{#1}}
\newcommand{\modelname}{SPICE\xspace}
\newcommand{\longmodelanme}{Self-supervised Person Image CrEation\xspace}
\newcommand\blfootnote[1]{%
  \begingroup
  \renewcommand\thefootnote{}\footnote{#1}%
  \addtocounter{footnote}{-1}%
  \endgroup
}

\iccvfinalcopy %

\ificcvfinal\fi

\creflabelformat{equation}{#2\textup{#1}#3}

\newenvironment{conditions*}
  {\par\vspace{\abovedisplayskip}\noindent
   \tabularx{\columnwidth}{>{$}l<{$} @{${}={}$} >{\raggedright\arraybackslash}X}}
  {\endtabularx\par\vspace{\belowdisplayskip}}

\begin{document}

\title{Learning Realistic Human Reposing using Cyclic Self-Supervision with 3D Shape, Pose, and Appearance Consistency}

 \author{Soubhik Sanyal$^{2*}$ \qquad Alex Vorobiov$^1$ \qquad Timo Bolkart$^2$ \qquad Matthew Loper$^1$ \\ \qquad Betty Mohler$^1$ \qquad Larry Davis$^1$ \qquad Javier Romero$^1$ \qquad Michael J. Black$^1$ \\
 \textrm{$^1$Amazon} \qquad \textrm{$^2$Max Planck Institute for Intelligent Systems, T\"ubingen}\\
{\tt\small \{ssanyal,tbolkart\}@tue.mpg.de} \quad
{\tt\small \{vorobioo,mloper,btmohler,lrrydav,javier,mjblack\}@amazon.com}
}

\input{notations}

\twocolumn[{%
	\renewcommand\twocolumn[1][]{#1}%
	\maketitle
	\begin{center}
		\newcommand{\teaserwidth}{\textwidth}
		\newcommand{\teaserheight}{0.75in}
		\centerline{
			\includegraphics[width=0.33\teaserwidth]{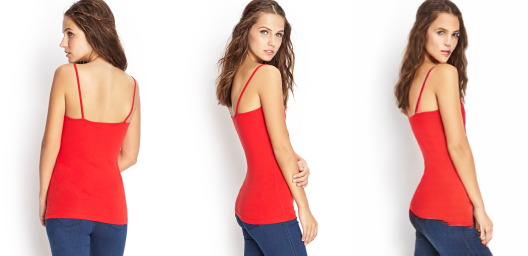} 
			\includegraphics[width=0.33\teaserwidth]{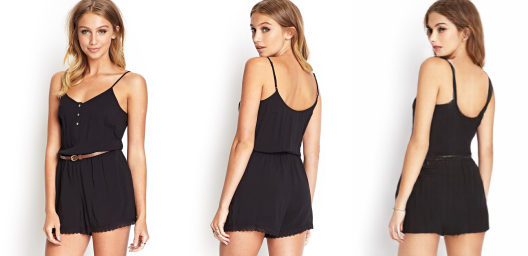} 
			\includegraphics[width=0.33\teaserwidth]{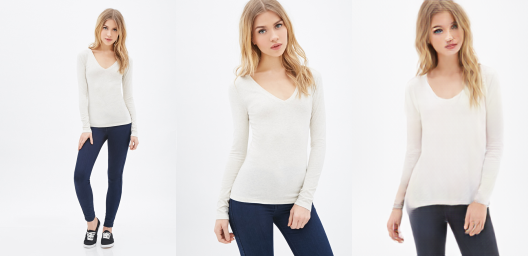} 
		}
		\centerline{
			\includegraphics[width=0.33\teaserwidth]{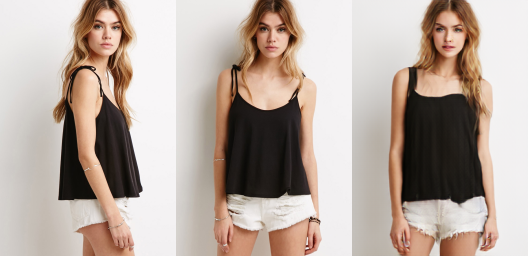} 
			\includegraphics[width=0.33\teaserwidth]{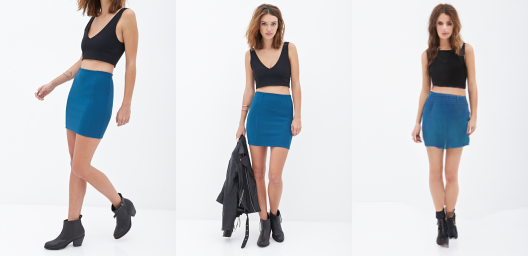} 
			\includegraphics[width=0.33\teaserwidth]{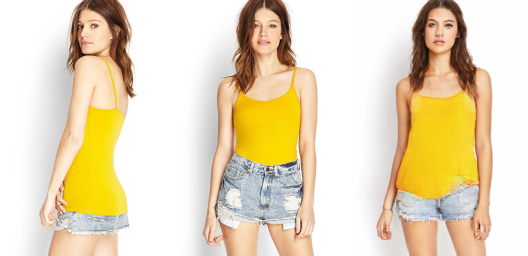} 
		}
	    \vspace{-0.1in}
		\captionof{figure}{{\bf \modelname} (\longmodelanme) generates an image of a person in a novel pose given a source image and a target pose. 
		Each triplet in the figure consists of the source image (left), a reference image in target pose (middle) and the generated image in the target pose (right); input and reference images are from the DeepFashion test set \cite{liuLQWTcvpr16DeepFashion}. 
		}
		\label{fig:teaser}
	\end{center}
	\vspace{-0.1in}
}]

\maketitle

\blfootnote{*This work was done during an internship at Amazon.}

\input{abstract}

\section{Introduction}

\input{introduction}

\section{Related Work}
\input{related_work}

\section{Method}
\input{proposed_method}

\section{Experiments}
\input{experiments}

\section{Conclusion}
\input{conclusion}

{\small
\balance

\input{egbib}
}

\newpage
\nobalance
\input{supmat}

\end{document}

%% file: notations.tex
\newcommand{\shapecoeff}{\boldsymbol{\beta}}
\newcommand{\shapedim}{{\left| \shapecoeff \right|}}
\newcommand{\shapespace}{\mathcal{S}}

\newcommand{\shapecoeffSource}{\shapecoeff_{s}}
\newcommand{\shapecoeffTarget}{\shapecoeff_{t}}
\newcommand{\shapecoeffSynth}{\hat{\shapecoeff}_{t}}

\newcommand{\posecoeff}{\boldsymbol{\theta}}
\newcommand{\posedim}{{\left| \posecoeff \right|}}
\newcommand{\posespace}{\mathcal{P}}

\newcommand{\posecoeffSource}{\posecoeff_{s}}
\newcommand{\posecoeffTarget}{\posecoeff_{t}}
\newcommand{\posecoeffSynth}{\hat{\posecoeff}_{t}}

\newcommand{\cameraParams}{\boldsymbol{\tau}}
\newcommand{\cameraParamsTarget}{\boldsymbol{\cameraParams_t}}
\newcommand{\cameraParamsSource}{\boldsymbol{\cameraParams_s}}
\newcommand{\cameraParamsSynth}{\boldsymbol{\hat{\cameraParams}_{t}}}
\newcommand{\cameraParamsSourceSynth}{\boldsymbol{\hat{\cameraParams}_s}}

\newcommand{\template}{\overline{\textbf{T}}}
\newcommand{\numjoints}{K}
\newcommand{\joints}{\textbf{J}}
\newcommand{\jointregressor}{\mathcal{J}}

\newcommand{\blendweights}{\mathcal{W}}
\newcommand{\blendweightsdim}{\left| \mathcal{W} \right|}
\newcommand{\numverts}{N}
\newcommand{\mesh}{\boldsymbol{\mathcal{M}}}

\newcommand{\fittingfn}{SMPLify}
\newcommand{\render}{Render}
\newcommand{\textureImage}{\mathcal{Q}}
\newcommand{\regressorSpin}{Spin}

\newcommand{\spice}{\modelname}
\newcommand{\generator}{\mathcal{G}}

\newcommand{\dataset}{\mathcal{D}}
\newcommand{\image}{I}
\newcommand{\pose}{P}
\newcommand{\rendering}{R}

\newcommand{\imageSource}{\image_s}
\newcommand{\poseSource}{\pose_s}
\newcommand{\renderingSource}{\rendering_s}

\newcommand{\imageTarget}{\image_t}
\newcommand{\poseTarget}{\pose_t}
\newcommand{\renderingTarget}{\rendering_t}

\newcommand{\imageSynth}{\hat{\image}_{t}}
\newcommand{\imageSourceSynth}{\hat{\image}_{s}}

\newcommand{\patch}{p}
\newcommand{\patchSource}{\patch_{s}}
\newcommand{\patchSynth}{\hat{\patch}_{t}}

\newcommand{\mask}{m}
\newcommand{\maskSource}{\mask_{s}}
\newcommand{\maskSynth}{\hat{\mask}_{t}}

\newcommand{\loss}{\mathcal{L}}
\newcommand{\lossIntermediate}{l}

\newcommand{\expectation}{\mathbb{E}}
\newcommand{\gram}{\mathbb{G}}

\newcommand{\relativePelvisRot}{\vert \posecoeffSource^{pel} - \posecoeffTarget^{pel} \vert}

\newcommand{\UNCOMPRESSED}{RAW}
\newcommand{\UncondD}{unconditional discriminator}

%% file: abstract.tex
\begin{abstract}
Synthesizing images of a person in novel poses from a single image is a highly ambiguous task.
Most existing approaches require paired training images; i.e.~images of the same person with the same clothing in different poses.
However, obtaining sufficiently large datasets with paired data is challenging and costly.
Previous methods that forego paired supervision lack realism.
We propose a self-supervised framework named \modelname (\longmodelanme) that closes the image quality gap with supervised methods.
The key insight enabling self-supervision is to exploit 3D information about the human body in several ways.
First, the 3D body shape must remain unchanged when reposing.
Second, representing body pose in 3D enables reasoning about self occlusions.
Third, 3D body parts that are visible before and after reposing, should have similar appearance features.
Once trained, \modelname takes an image of a person and generates a new image of that person in a new target pose.
\modelname achieves state-of-the-art performance on the DeepFashion dataset, improving the FID score from 29.9 to 7.8 compared with previous unsupervised methods, and with performance similar to the state-of-the-art supervised method (6.4).
\modelname also generates temporally coherent videos given an input image and a sequence of poses, despite being trained on static images only.
\end{abstract}

%% file: introduction.tex
Given a single source image of a person, can we generate a realistic image of what they would look like from a different viewpoint, in a different pose? 
While this problem is inherently ambiguous, there is significant statistical regularity in human pose, clothing, and appearance, that could make this possible as illustrated in Fig.~\ref{fig:teaser}.
A solution to the problem would have widespread applications in online fashion, gaming, personal avatar creation or animation, and has consequently generated significant research interest \cite{Dong_2020_CVPR, knoche2020reposing, Ren_2020_CVPR, shiloss, song2019unsupervised, yang2020towards}.

Recent work focuses on generative modeling \cite{goodfellow2014generative, isola2017image, karras2019style, CycleGAN2017}, especially using conditional image synthesis. 
One set of methods uses supervised training \cite{Dong_2020_CVPR, li2019dense, liu2019liquid, Sarkar2020}, which requires paired training images of the same person in different poses with the same appearance and clothing. %
Requiring such paired data limits the potential size of the training set, which can impair robustness and generalization.
Consequently, we address this problem {\em without any paired data} by developing a self-supervised approach.
Such self-supervised formulations have also received significant recent attention
\cite{esser2018variational, ma2018disentangled, pumarola2018unsupervised, yang2020towards}.  

Our novel formulation builds on the idea of cycle-consistency \cite{CycleGAN2017} with some important modifications.
For the forward direction of the cycle, the method takes a source image, source pose and target pose and generates a target image conditioned on pose and appearance. 
The reverse direction takes this generated image and regenerates the source image by switching the source and target conditions.
The goal is to minimize the difference between the original input image and the one synthesized through the cycle. %
The problem is that this approach can have a trivial solution in which the cycle produces the identity mapping.
To address this, previous methods
\cite{pumarola2018unsupervised, song2019unsupervised} constrain the target image generation with 2D information.
Human bodies, however, are non-rigid 3D entities and their deformations and occlusions are not easily expressed in 2D.
We show how leveraging 3D information, automatically extracted from images, constrains the model in multiple ways.

\begin{figure}[t]
	\centerline{
		\includegraphics[width=1.0\linewidth]{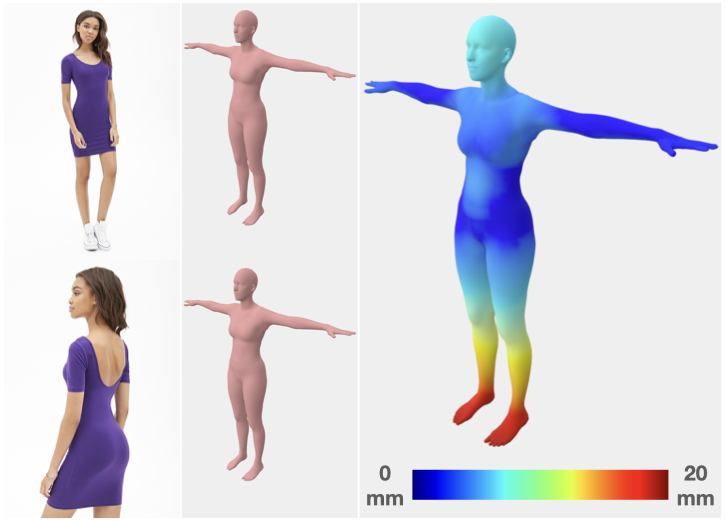}
	}
	\caption{\textbf{Shape consistency:} 
	The first column shows two images of the same person in two different poses and views. 
	The second column shows the 3D bodies predicted by our 3D regressor and posed in a T-pose.
	The estimated 3D body shape is similar for the same subject across poses and views. 
	The third column shows the per-vertex difference of both meshes, color coded from blue (0 mm) to red ($20$ mm).
	}
\vspace{-0.1in}
	\label{fig:shapeconsistency}
\end{figure}

Specifically, our method, called \modelname (\longmodelanme), exploits the estimated 3D body to constrain the image generation, enabling self-supervised learning. 
In particular, we estimate the SMPL body model \cite{SMPL:2015} parameters corresponding to both the input and the generated target image.
Since the input and target image only differ in terms of their pose, their body shape should be the same. 
SMPL makes this easy to enforce because it factors body shape from pose.
Using this we introduce two losses.
First, we use a pose loss that encourages the body pose in the generated image to match the target pose in 3D. 
Second, we add a shape consistency loss that encourages the person in the generated image to have the same 3D shape as the person in the source image (Fig.~\ref{fig:shapeconsistency}).

\begin{figure}[t]
	\centerline{
		\includegraphics[width=0.7\linewidth]{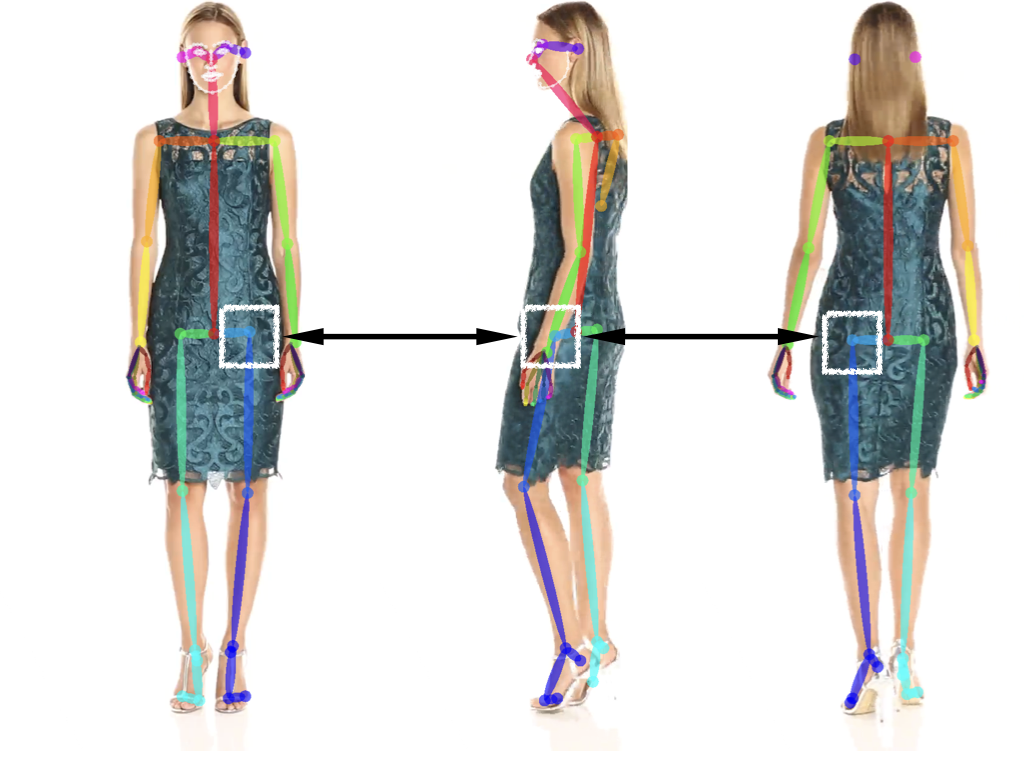}
	}
	\caption{\textbf{Problem: patch loss based on 2D keypoints.} A person is seen in three different poses with the same clothing. A patch (white rectangle) is extracted at her left hip keypoint.  
	Assuming that the appearance of the patch is the same across viewpoints is incorrect.
	Instead, \modelname uses the 3D body surface to reason about the regions of the body that are visible in multiple views.
	Keypoints are predicted by OpenPose \cite{8765346} for this figure.}
\vspace{-0.3in}
	\label{fig:flawsof2Dpatchloss}
\end{figure}

These two constraints, however, are not sufficient to generate images with the correct appearance since they only force the model to generate an image with the right shape and pose.
There is no constraint that the generated image has the appearance of the source image (e.g.~clothing, hair, etc.).
Prior work addresses this by enforcing a perceptual loss between patches at each 2D joint \cite{pumarola2018unsupervised}.
This is not sufficient when the body is seen with large viewpoint changes or
where a body part becomes occluded; see Fig.~\ref{fig:flawsof2Dpatchloss}.
We solve this problem by introducing pose-dependent appearance consistency on the body surface instead of at the joints. 
The idea is that the projected surface of the 3D body in two different poses must have  similar appearance features for matching parts of the body and this similarity should be weighted proportional to the relative global orientation difference between the 3D bodies.

In summary, we improve the realism of self-supervised human reposing by exploiting 3D body information in three novel ways: using a 3D pose loss, body shape consistency, and occlusion-aware appearance feature consistency.
We train \modelname with our new constraints on unpaired data. 
Experiments on the DeepFashion  \cite{liuLQWTcvpr16DeepFashion} and Fashion Video datasets \cite{zablotskaia2019dwnet} show the effectiveness of \modelname qualitatively and quantitatively.
\modelname significantly outperforms the prior state-of-the-art (SOTA) un/self-supervised methods and is nearly as accurate as the best supervised methods.

%% file: related_work.tex
\begin{figure*}[ht]
	\centerline{
		\includegraphics[width=1.0\linewidth]{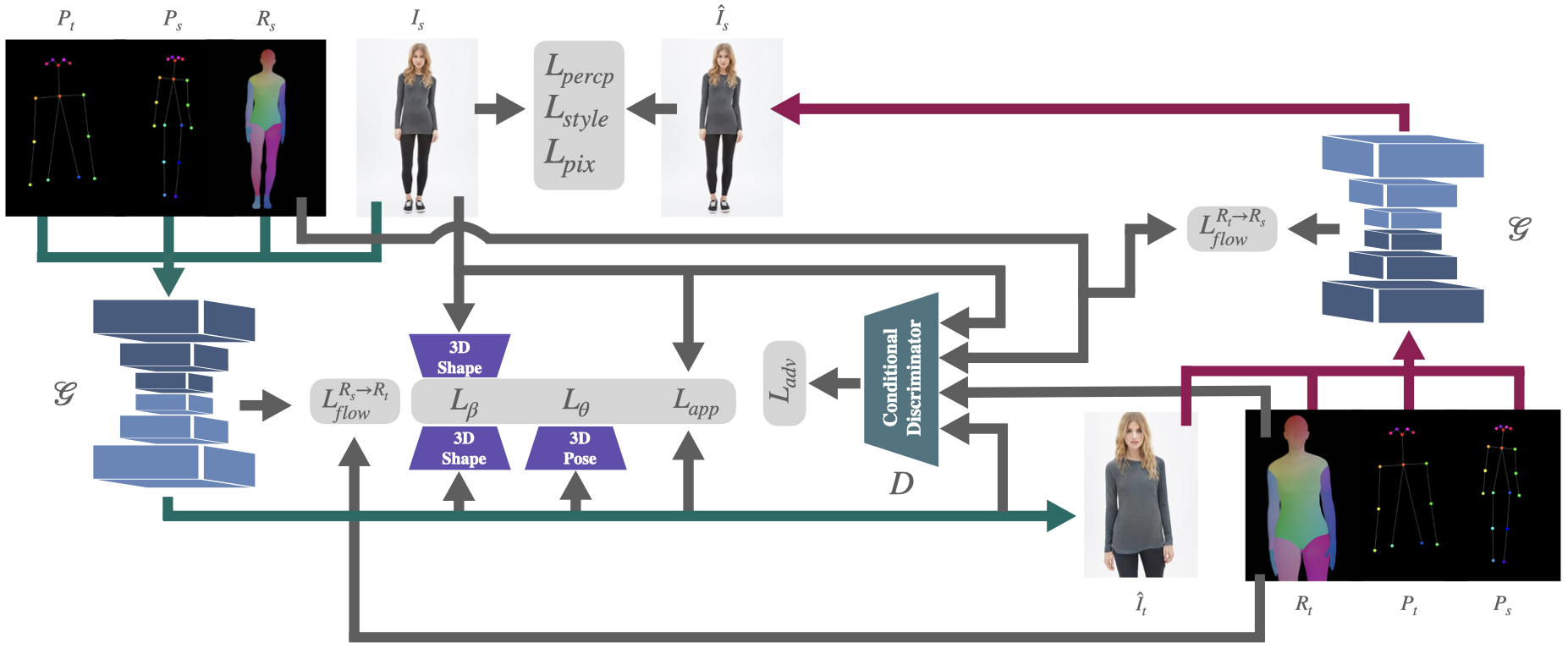}
	}
	\vspace{-0.1in}
	\caption{\textbf{Overview of \modelname:} 
	Given a source image of a person $\imageSource$, source pose $\poseSource$, target pose $\poseTarget$ and 3D mesh rendering of the source pose $\renderingSource$, the generator $\generator$ generates a target image with the person in the target pose. 
	Then the source and target pose are swapped and passed through $\generator$ but with the generated target image  as the source. 
	This should re-generate the source image enabling the use of a cyclic self-supervision loss, $\loss_{cycle}$, during training. 
	To prevent trivial solutions, the cycle is constrained by losses on 3D pose $\loss_{\posecoeff}$, shape $\loss_{\shapecoeff}$ and appearance $\loss_{app}$, which are the main contributions of \modelname (\cref{proposed_method}),  and an adversarial loss $\loss_{adv}$.
	Note that the $\poseSource$ and $\poseTarget$ are provided as input heat-maps to $\generator$.
	}
	\label{fig:overview}
	\vspace{-0.1in}
\end{figure*}

Methods for reposing images of humans can be broadly divided into two categories: supervised or unsupervised. 
While both approaches rely on generative modeling \cite{goodfellow2014generative, isola2017image, karras2019style, CycleGAN2017}, the 
supervised methods require paired ground truth: source and target training images of the person in different poses.
Our approach falls into the unsupervised or self-supervised category, in which we do not use paired training data.
We address each class of methods below.

\qheading{Supervised methods:}
Supervised approaches learn to transform a source image given a source pose and a target pose  \cite{balakrishnan2018synthesizing, dong2018soft, Dong_2020_CVPR, grigorev2019coordinate, knoche2020reposing, li2019dense, liu2019liquid, ma2017pose, ma2020unselfie, men2020controllable, neverova2018dense, Ren_2020_CVPR, Sarkar2020, Siarohin_2019_PAMI, siarohin2018deformable, Siarohin_2018_CVPR, tang2020xinggan, tang2021structure, yang2018pose, yang2020towards2d, zanfir2020human, zhu2019progressive, lakhal2018pose}. 
Supervision is provided by the target image during training and usually adversarial and perceptual losses are used to train the model \cite{shiloss}. 
The differences between methods usually lie in network inputs and their architectures.
Dong et al.~\cite{dong2018soft} synthesize the target image in two stages. First they generate a target pose segmentation from the source pose and use it in their soft-gated warping block architecture to render the person in the target pose. 
Knoche et al.~\cite{knoche2020reposing} learn an implicit volumetric representation of the person to warp the source pose into the target pose. The volumetric representation is implicitly learned using an encoder decoder architecture.
Li et al.~\cite{li2019dense} utilize a learned flow field to warp a person in a source pose to the target pose. The flow field is learned from 3D bodies and is used for warping at the feature level and pixel level in a deep architecture.
Ma et al.~\cite{ma2017pose} first generate a coarse image of the global structure of a human in the target pose from the source pose in a two stage network. This is then refined in an adversarial way in the second stage to get finer details.
Sarkar et al.~\cite{Sarkar2020} compute a partial UV texture map using DensePose \cite{guler2018densepose} from the source image. They use this as input to their network, which learns to complete the UV texture map and render it in a target pose using neural rendering.
Siarohin et al.~\cite{siarohin2018deformable} propose a network architecture using deformable skip connections to tackle the problem. 
Tang et al.~\cite{tang2020xinggan} propose a co-attention fusion model that fuses appearance and shape features from images, which they disentangle inside their architecture. They use two different discriminators for appearance and shape to jointly judge the generation.
Zhu et al.~\cite{zhu2019progressive} propose a progressive generator using a sequence of attention transfer blocks. Each of these blocks transfers certain regions it attends to and generates the image of the person progressively.
Ren et al.~\cite{Ren_2020_CVPR} propose a new deep architecture where they combine flow-based operations with an attention mechanism. 
Note that the above methods are all supervised and cannot be directly used in the self-supervised scenario. In contrast, our work is focused on unpaired data and we build on the Ren et al.~\cite{Ren_2020_CVPR} architecture to enable this. Thus our contribution does not lie with network architecture but, rather, in introducing novel constraints that make it possible to solve the problem without paired data.

\qheading{Unsupervised or self-supervised settings: }
There is an increasing interest in solving the problem in an unsupervised/unpaired manner.
Such approaches can work when paired data is not available or can increase robustness and generalization by combining paired and unpaired data.
An early approach \cite{ma2018disentangled} divides the process in two stages. 
The first stage uses an auto-encoder-based architecture to learn the corresponding embedding space for pose, foreground and background from source images.
The second stage maps Gaussian noise to the embedding space of pose, foreground and background and uses the pretrained decoder from the first stage to generate a person's image in a new pose.
Yang et al.~\cite{yang2020towards} train an appearance encoder from the source image to learn the appearance representation or embedding. They fuse the appearance embedding with the pose embedding coming from an image of a different person in different pose. In this way they generate the person's image in the new pose.
Esser et al.~\cite{esser2018variational} use a U-Net architecture conditioned on the output of a variational auto-encoder for appearance. They also try to disentangle pose and appearance of a person from the source images.

In general, the mentioned approaches attempt to disentangle shape, pose and appearance in the latent space from a 2D image, which is a hard problem.
This results in a relatively poor image generation quality.
Instead of learning this disentanglement from images we approach the problem differently.
We extract the person's pose and shape information in a parametric decoupled 3D body representation, SMPL \cite{SMPL:2015}, and constrain our self-supervised generation.
Furthermore, we also constrain appearance generation by leveraging the surface and projection of the 3D body.

Similar to our cyclic formulation, Pumarola et al.~\cite{pumarola2018unsupervised} and Song et al.~\cite{song2019unsupervised} train their networks in a self-supervised CycleGAN \cite{CycleGAN2017} fashion.
Additionally, \cite{song2019unsupervised}  use semantic parsing maps as input to the network.
They constrain their self-supervised generation with 2D information.
We differ from these methods by constraining the self-supervised approach with 3D body information.

%% file: proposed_method.tex
\label{proposed_method}

\modelname requires a training dataset of tuples $(\image, \pose, \rendering)$, each containing an image $\image$ of a person, their pose $\pose$ as 2D keypoints, and a 2D rendering $\rendering$.
To generate $\rendering$ we fit the SMPL 3D mesh \cite{SMPL:2015} to  $\pose$ using SMPLify~\cite{bogo2016SMPLify}, and render 
the mesh using a color wheel texture in UV space.

We treat all the samples in the dataset as independent; that is, our method does not require images of the same person wearing the same clothing in different poses (i.e.~without direct supervision through paired data).
During training, the source image $\imageSource$, source pose $\poseSource$, source rendering $\renderingSource$, target pose $\poseTarget$ and target rendering $\renderingTarget$ are given. 
\modelname then synthesizes the image $\imageSynth$, which is the reposed source image $\imageSource$, using a generator network $\generator$ (\cref{sec:generator}):
\begin{equation}
    \imageSynth = \generator(\imageSource, \poseSource, \renderingSource, \poseTarget).
    \label{eq:generation_of_synth_image}
\end{equation}
During training, we exploit cycle-consistency (\cref{sec:closing_cycle}).  
Specifically, we generate a synthetic version of the source image from $\imageSynth$ by reusing $\generator$; i.e.~$\imageSourceSynth = \generator(\imageSynth,\poseTarget,\renderingTarget,\poseSource)$. 
This enables us to directly apply perceptual and pixel-wise losses between $\imageSource$ and $\imageSourceSynth$ to train $\generator$.
To prevent trivial solutions, we add 3D guidance (\cref{sec:shape_and_pose_consistency}) and appearance constraints (\cref{sec:appearence_consistency}) for $\imageSynth$.
See \cref{fig:overview} for an overview of the \modelname training pipeline.

\subsection{Generator architecture}
\label{sec:generator}

Our generator $\generator$ has two modules: a global flow field estimator and a local neural rendering module.
The flow estimator module takes $\renderingSource$, $\poseSource$, $\poseTarget$ as input and generates 2D warping fields at the feature level between the source and the target pose.
The neural rendering module takes $\imageSource$ and $\poseTarget$ as inputs and uses the generated warping fields at the feature level of its local attention blocks to generate $\imageSynth$.
The loss for the flow estimator module can be written as,
\begin{equation}
    \label{eq:flow_loss}
    \loss_{flow} = \loss_{flow}^{\renderingSource \rightarrow \renderingTarget} + \loss_{flow}^{\renderingTarget \rightarrow \renderingSource},
\end{equation}
where $\loss_{flow}^{x\rightarrow y}$ is the weighted addition of the sampling correctness loss and regularization loss for the generated flow fields, as proposed by Ren et al.~\cite{Ren_2020_CVPR}.
Here, $\loss_{flow}^{\renderingSource \rightarrow \renderingTarget}$ is applied while synthesizing $\imageSynth$, and $\loss_{flow}^{\renderingSource \rightarrow \renderingTarget}$ is applied when regenerating $\imageSourceSynth$ at the end of the cycle.
The sampling correctness loss is the computed cosine similarity distance between the warped source features and target features.
The source and target features come from a specific layer of a pre-trained VGG network~\cite{simonyan2014very} given the source and target renderings as input, respectively. 
The regularisation loss provides regularisation to the generated warping fields.

Our generator follows the design of Ren et al.~\cite{Ren_2020_CVPR} with the difference that the flow estimator is trained on source and target renderings (i.e. $\renderingSource$ and $\renderingTarget$), instead of source and target images (i.e. $\imageSource$ and $\imageTarget$), due to the unavailability of $\imageTarget$ in our setting.
We refer to Ren et al.~\cite{Ren_2020_CVPR} for more details on the sampling correctness loss and the regularization loss.

\subsection{Closing the cycle}
\label{sec:closing_cycle}
Enforcing cycle consistency enables us to train \modelname with supervised losses between the source image $\imageSource$ and the regenerated source image $\imageSourceSynth$.
Specifically, we minimize
\begin{equation}
    \label{eq:regen_source_loss}
    \loss_{cycle} = \lambda_{percep} \loss_{percep} + \lambda_{style} \loss_{style} + \lambda_{pix} \loss_{pix},
\end{equation}
where the $\lambda$'s are individual loss weights, and the perceptual loss $\loss_{percep}$, style loss $\loss_{style}$, \cite{johnson2016perceptual}, and pixel-wise loss $\loss_{pix}$ are defined as
\begin{gather*}
    \loss_{percep} =  \sum _j \left \| \phi_j(\imageSource) - \phi_j(\imageSourceSynth) \right \|_1 \\
	\loss_{style} = \sum _j \left \| \mathbb{G}(\phi_j(\imageSource)) - \mathbb{G}(\phi_j(\imageSourceSynth)) \right \|_1 \\
	\loss_{pix} = \left \| \imageSource - \imageSourceSynth \right\|_1,
\end{gather*}
where $\phi_j$ is the activation map of the $j^{th}$ layer of a pretrained VGG network \cite{simonyan2014very}, and $\gram$ is the Gram matrix built from the activation map $\phi_j$.

To generate realistic looking images, \modelname minimizes an adversarial loss by adding a discriminator, $D$, that discriminates between the fake images $\imageSynth$ and real images $\imageSource$. 
To provide pose information along with each image, we condition $D$ on the corresponding rendering (i.e.~$\renderingTarget$ for $\imageSynth$, and $\renderingSource$ for $\imageSource$), by providing the concatenation of the two images as discriminator input.
Formally, we minimize%
\begin{equation}
    \loss_{adv} = \expectation[\log (1 - D(\imageSynth, \renderingTarget))] + \expectation[\log D(\imageSource, \renderingSource)].
\end{equation}

\subsection{Pose and shape consistency}
\label{sec:shape_and_pose_consistency}

\modelname uses the SMPL~\cite{SMPL:2015} 3D human body model to enforce pose and shape consistency during training.
SMPL combines identity-dependent shape blendshapes with pose-dependent corrective blendshapes and linear blendskinning (LBS) for pose articulation. 
Importantly, this formulation disentangles body shape form pose.
Given parameters for shape $\shapecoeff \in \mathbb{R}^\shapedim$ and pose $\posecoeff \in \mathbb{R}^{3\numjoints+3}$, SMPL is a function, $\mesh(\shapecoeff, \posecoeff)$ that outputs a 3D mesh with $\numverts=6890$ vertices.

To extract SMPL shape and pose parameters $\shapecoeff$ and $\posecoeff$ from $\image$, we use a differentiable regressor \cite{kolotouros2019spin}, denoted as
\begin{equation}
    \shapecoeff, \posecoeff = f_{3D}(\image).
    \label{eq:shape_pose_regressor}
\end{equation}
Given the extracted SMPL parameters $\shapecoeffSynth, \posecoeffSynth = f_{3D}(\imageSynth)$, we define a loss that encourages the 3D rotation of the joints in the synthetic image, $\posecoeffSynth$, to be the same as the rotation of the joints in the target pose $\posecoeffTarget$:
\begin{equation}
    \label{eq:3D_jointloss}
    \loss_{\posecoeff} = \left \| \posecoeffTarget - \posecoeffSynth \right \|_1,
\end{equation}
where $\posecoeffTarget$ is obtained by running SMPLify~\cite{bogo2016SMPLify} on $\poseTarget$.

\modelname also enforces body shape consistency (\cref{fig:shapeconsistency}) based on the observation that while $\imageSource$ and $\imageSynth$ differ in pose, their body shapes $\shapecoeffSource$ (i.e.~$\shapecoeffSource, \posecoeff_s = f_{3D}(\imageSource)$) and $\shapecoeffSynth$ must be the same, enforced by
\begin{equation}
    \label{eq:shape_consistency}
    \loss_{\shapecoeff} = \left \| \shapecoeffSource - \shapecoeffSynth \right \|_1.
\end{equation}

\subsection{Appearance feature consistency}
\label{sec:appearence_consistency}

\begin{figure}[t]
	\centerline{
	    \includegraphics[width=0.35\columnwidth]{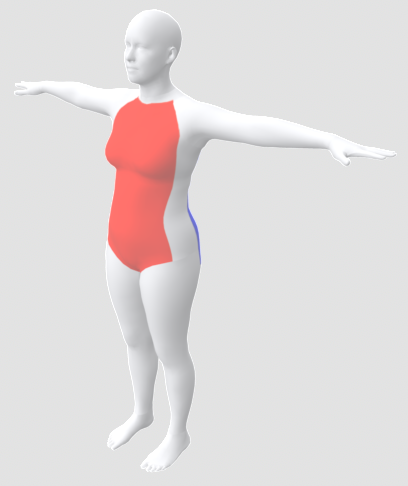}
		\includegraphics[width=0.12\columnwidth,trim={10.8cm, 0cm, 10.8cm, 0cm}, clip]{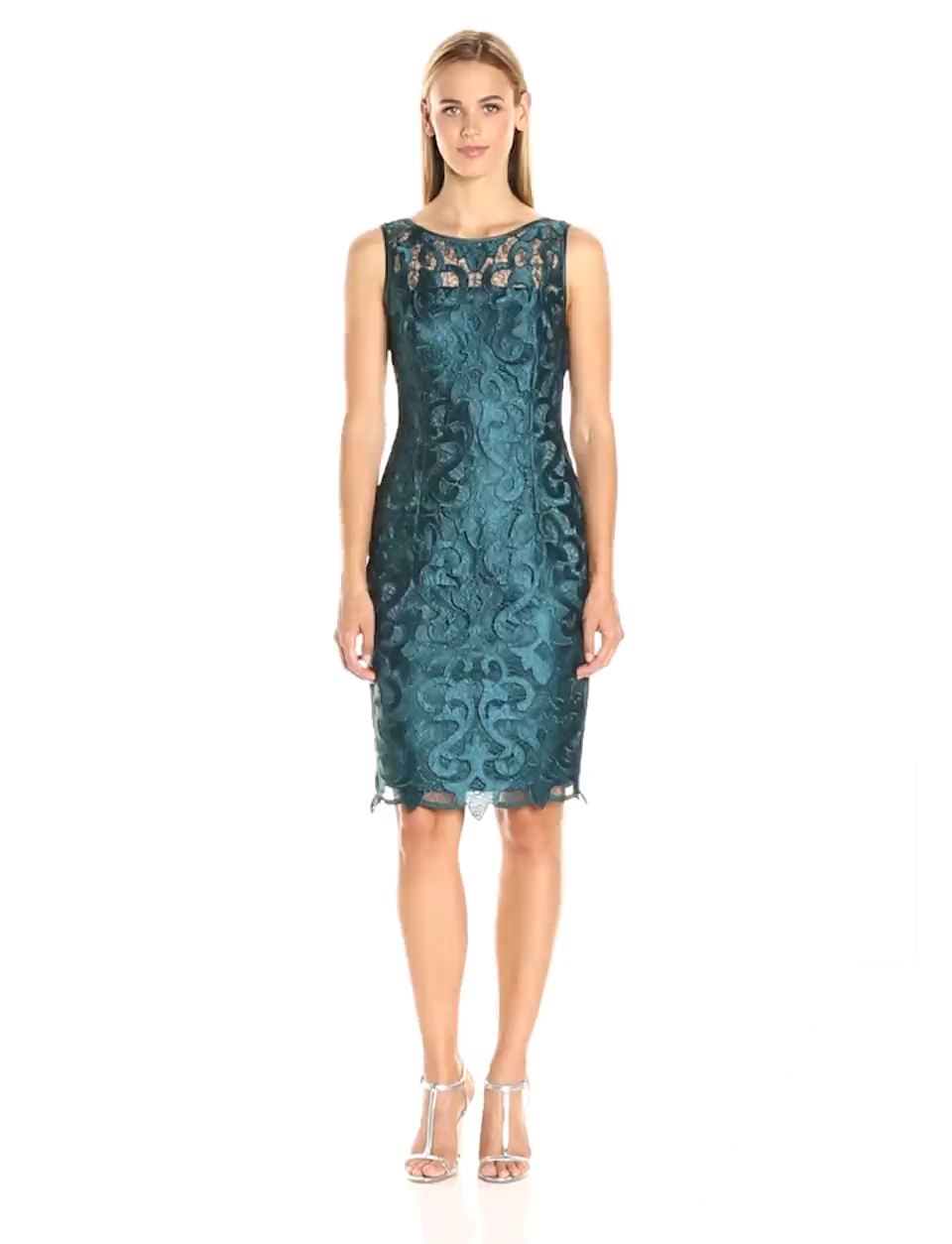}
		\includegraphics[width=0.12\columnwidth,trim={10.8cm, 0cm, 10.8cm, 0cm}, clip]{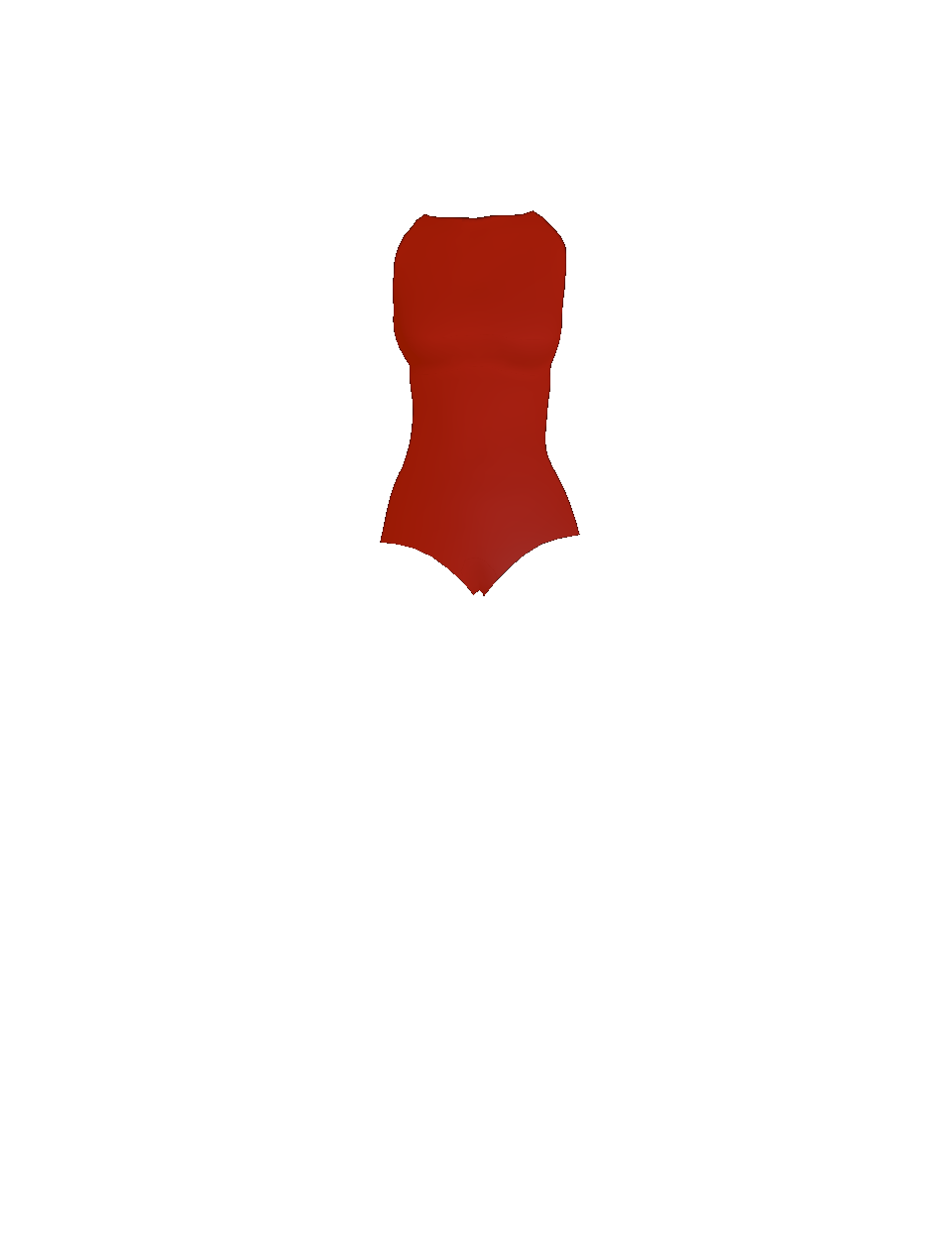}
        \includegraphics[width=0.12\columnwidth,trim={10.8cm, 0cm, 10.8cm, 0cm}, clip]{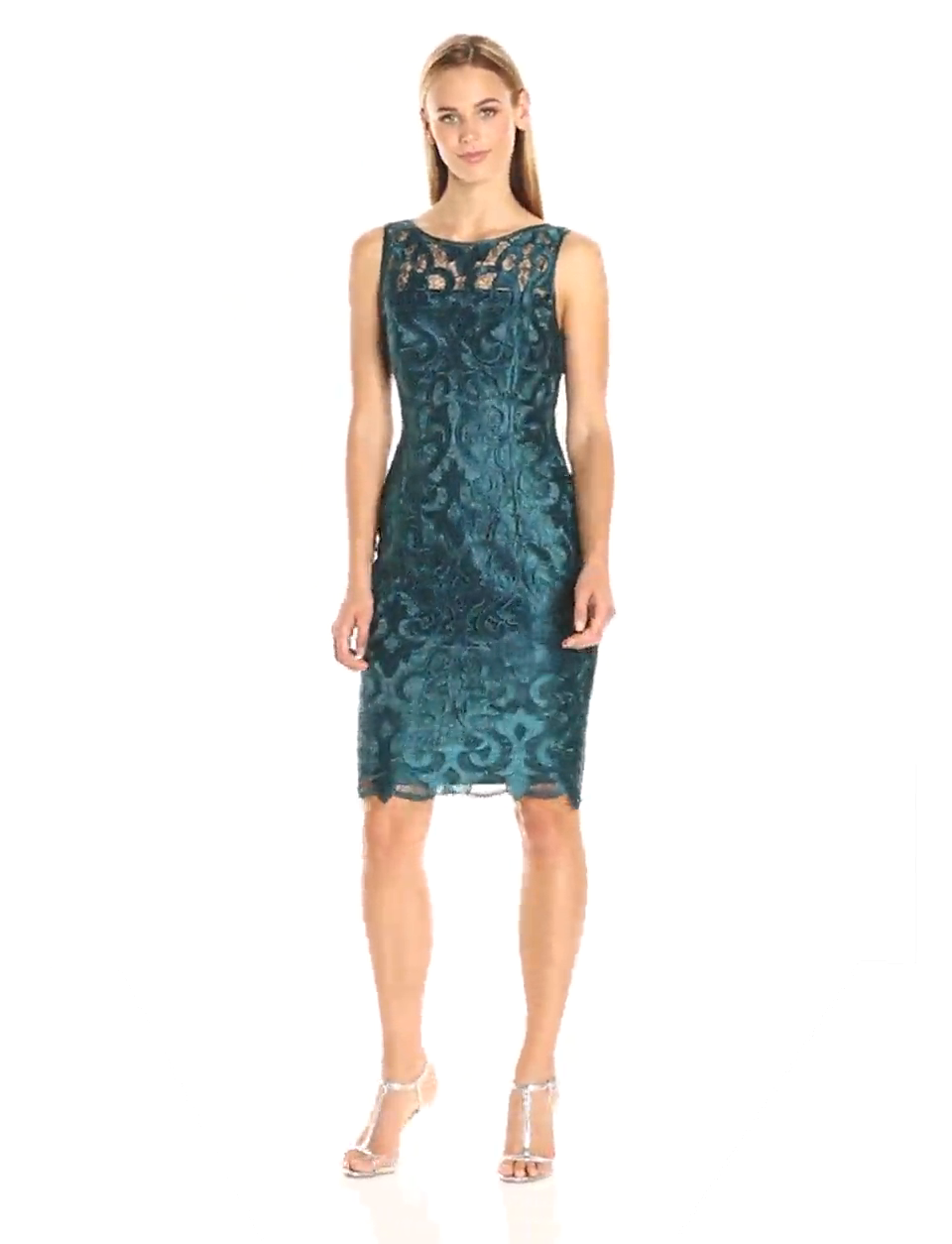}
        \includegraphics[width=0.12\columnwidth,trim={10.8cm, 0cm, 10.8cm, 0cm}, clip]{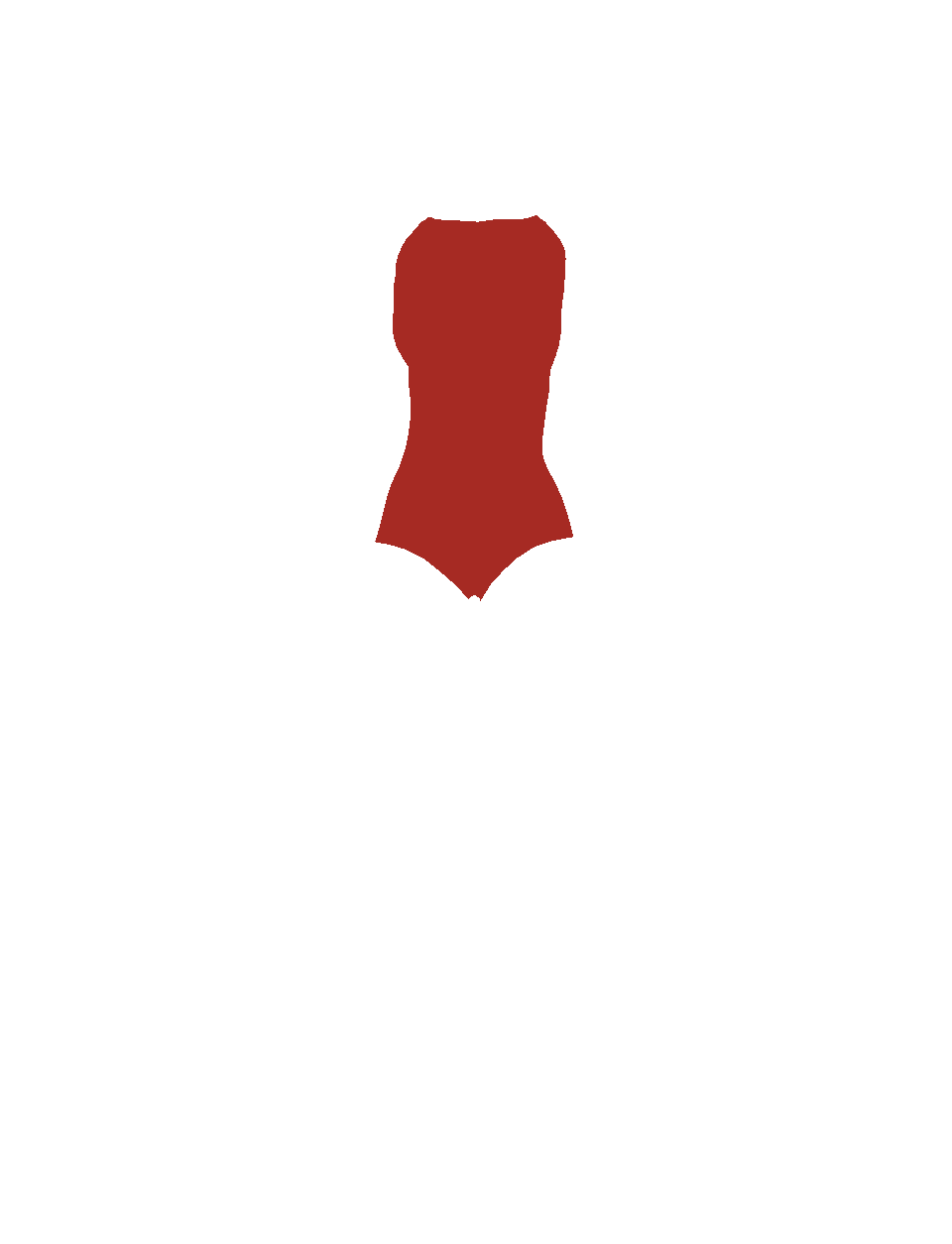}
	}
	\centerline{
	    \includegraphics[width=0.35\columnwidth]{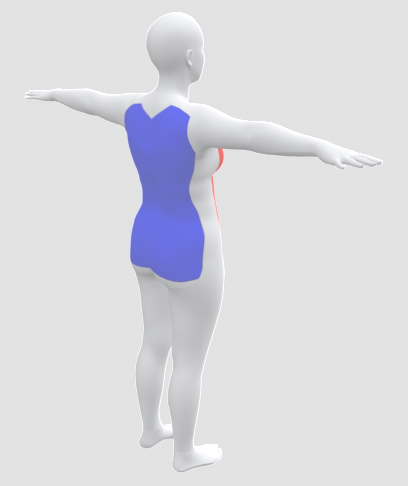}
		\includegraphics[width=0.12\columnwidth,trim={10.8cm, 0cm, 10.8cm, 0cm}, clip]{images/surface_patch_loss/raw_images/91+20mY7UJS_020.png}
		\includegraphics[width=0.12\columnwidth,trim={10.8cm, 0cm, 10.8cm, 0cm}, clip]{images/surface_patch_loss/raw_images/91+20mY7UJS_020_frontbodytorso_whitebackground.png}
		\includegraphics[width=0.12\columnwidth,trim={10.8cm, 0cm, 10.8cm, 0cm}, clip]{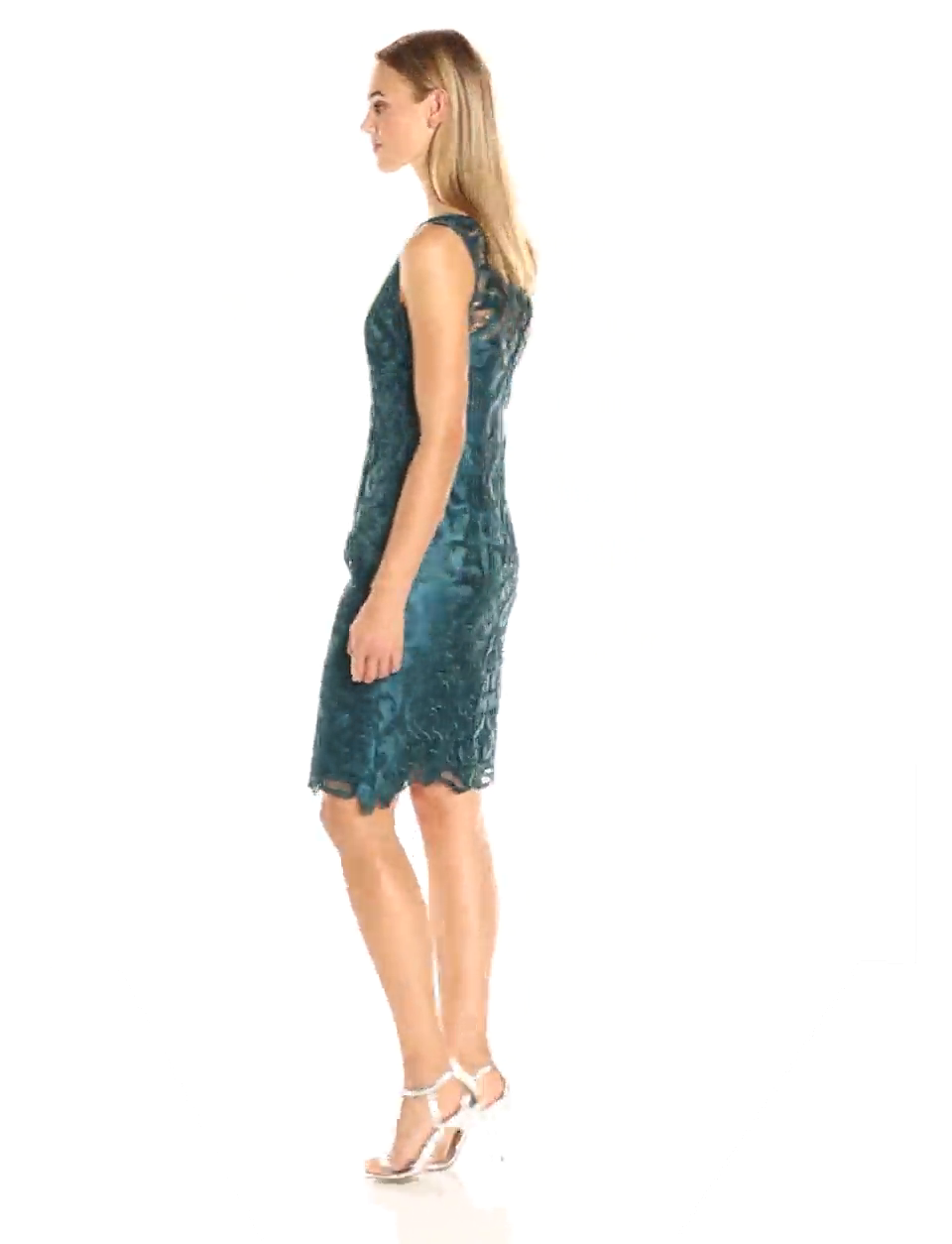}
		\includegraphics[width=0.12\columnwidth,trim={10.8cm, 0cm, 10.8cm, 0cm}, clip]{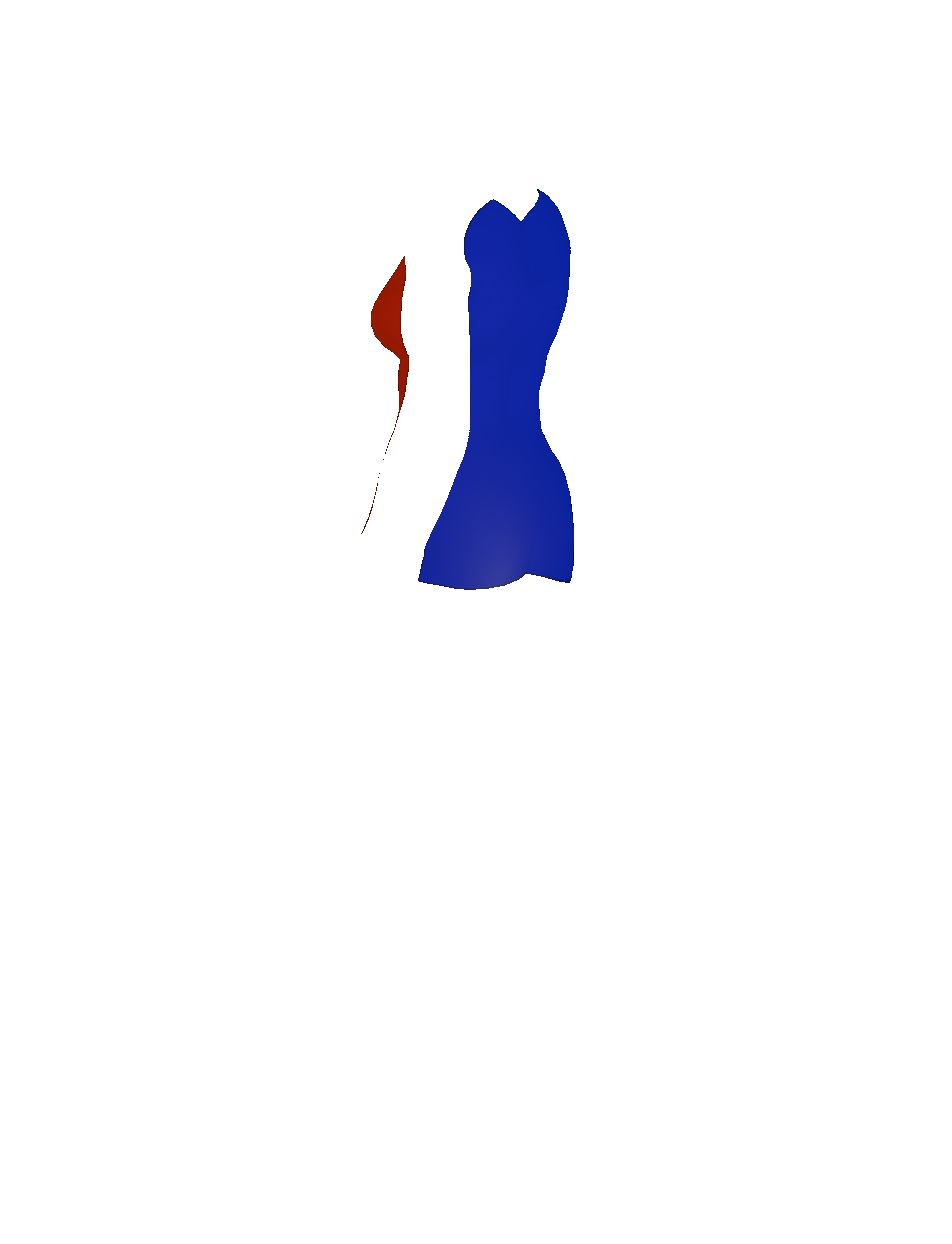}
	}
	{
    \hspace{0.18\columnwidth} a) \hspace{0.30\columnwidth} b) \hspace{0.20\columnwidth} c) %
	}
	\caption{
	\textbf{Appearance feature consistency:}
	a) SMPL template with front (red) and back (blue) torso masks, 
	b) and c) shows images of a person in different poses (left), and corresponding torso masks obtained by rendering the 3D body with the subject's pose.
	The appearance consistency loss is then applied on image segments for torso masks of the same color weighted by the relative pelvis rotation.
	}
	\label{fig:bodyfrontbacktexture}
\end{figure}

The above losses constrain pose and shape in $\imageSynth$, but do not guarantee that the appearance of $\imageSynth$ remains consistent with $\imageSource$.
Consequently, we formulate an additional constraint on the appearance of matching regions in $\imageSource$ and $\imageSynth$ to be similar.
Due to the unconstrained change in the pose between $\imageSource$ and $\imageSynth$, we cannot apply the deep appearance loss (perceptual or style loss) directly between those images.

Instead, we leverage the 3D body mesh to apply an appearance loss between corresponding image segments.
Given SMPL parameters $\shapecoeff$ and $\posecoeff$, we render the mesh $\mesh(\shapecoeff, \posecoeff)$ with the texture from \cref{fig:bodyfrontbacktexture}~{a)} to get the image segments for the rendered front and back torso areas, shown in \cref{fig:bodyfrontbacktexture}~{b), and c)}.
Let $M_{mask}$ denote a binary mask, with value $1$ for pixels within the front/back torso segment and $0$ elsewhere.
Further, let $\mathcal{P}_{patch}$ denote $\image \odot M_{mask}$, where $\odot$ is the Hadamard product. 
Both, $\mask$ and $\patch$ are cropped from $M_{mask}$ and $\mathcal{P}_{patch}$, by the bounding box of the image segment.

Given image patches $\patchSource$ and $\patchSynth$ together with binary masks $\maskSource$ and $\maskSynth$, both extracted from $\imageSource$ and $\imageSynth$, the appearance consistency is given as $\lossIntermediate_{app} =$:
\begin{equation}
    \begin{split}
    \lambda_{a_1} \sum _k \left \| \phi_k(\patchSource) \odot \psi_k(\maskSource) - \phi_k(\patchSynth) \odot \psi_k(\maskSynth) \right \|_1 \\
                + \lambda_{a_2} \sum _j \left \| \gram_j^{\phi}(\patchSource) \odot \psi_j(\maskSource)  - \gram_j^{\phi}(\patchSynth) \odot \psi_j(\maskSynth) \right \|_1
    \end{split}
\end{equation}
where
the $\lambda$'s are weights, $\phi_k$ is the activation map of the $k^{th}$ layer of pretrained VGG network \cite{simonyan2014very}, $\mathbb{G}_j^{\phi}$ is the Gram matrix built from the corresponding activation map $\phi$, and $\psi$ is the down-sampling function for the corresponding layer.

Note that the appearance loss as it is formulated requires sufficient overlap of corresponding image features within the mask crop.  
We compute the appearance loss as:
\begin{equation}
    \loss_{app} = \lambda(\posecoeffSource,\posecoeffTarget) \times \lossIntermediate_{app},
\end{equation}
where $\lambda(\posecoeffSource,\posecoeffTarget)$ is a weighting function that depends on the relative pelvis rotation (i.e.~rotation around the SMPL root joint) between the source and target pose:
\begin{equation}
  \lambda(\posecoeffSource,\posecoeffTarget)=\left\{
  \begin{array}{@{}ll@{}ll@{}ll@{}}
    1.0 & \text{if}\ 0^\circ \leq \relativePelvisRot < 20^\circ \\
    0.1 & \text{if}\ 20^\circ \leq \relativePelvisRot < 40^\circ \\
    0.01 & \text{if}\ 40^\circ \leq \relativePelvisRot < 60^\circ \\
    0.0 & \text{otherwise} .
  \end{array}\right.
\end{equation}

\subsection{Final loss}
\label{sec:final_loss}
The total loss of the proposed approach is:
\begin{equation}
\begin{split}
    \loss_{\spice} = & \alpha_{cycle} \loss_{cycle} + \alpha_{flow} \loss_{flow} + \alpha_{adv} \loss_{adv} \\
    & + \alpha_{\posecoeff} \loss_{\posecoeff} + \alpha_{\shapecoeff} \loss_{\shapecoeff} + \alpha_{app} \loss_{app},
\end{split}
\end{equation}
where  $\alpha_{i}$ are the corresponding loss weights. 
The following section provides details on how these weights are set.

%% file: experiments.tex
\label{sec:experiments}

\begin{figure}
    \centerline{
        \includegraphics[width=0.14\linewidth, height=2.5cm,trim={2.0cm, 0cm, 1.5cm, 0cm}, clip]{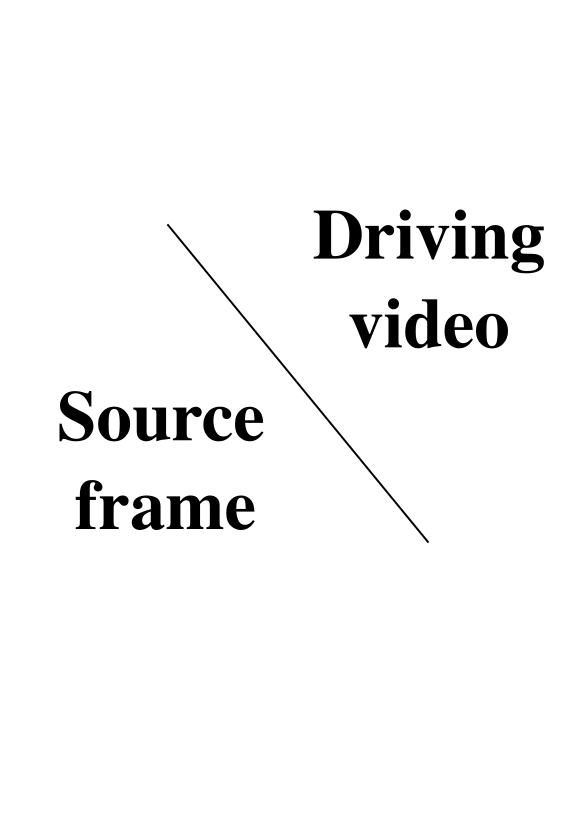} 
		\includegraphics[width=0.17\linewidth,trim={1.5cm, 0cm, 1.5cm, 0cm}, clip]{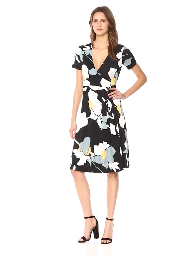}
		\includegraphics[width=0.17\linewidth,trim={1.5cm, 0cm, 1.5cm, 0cm}, clip]{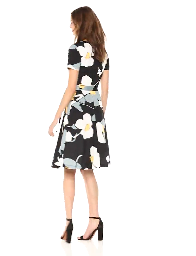}
		\includegraphics[width=0.17\linewidth,trim={0.8cm, 0cm, 2.2cm, 0cm}, clip]{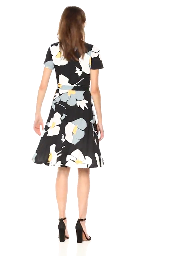}
		\includegraphics[width=0.17\linewidth,trim={0.8cm, 0cm, 2.2cm, 0cm}, clip]{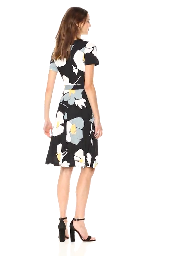}
		\includegraphics[width=0.17\linewidth,trim={1.8cm, 0cm, 1.2cm, 0cm}, clip]{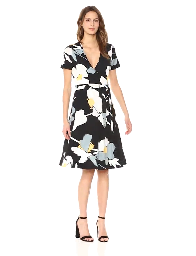}
	}
	\centerline{
        \includegraphics[width=0.17\linewidth,trim={1.5cm, 0cm, 1.5cm, 0cm}, clip]{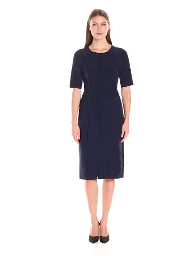} 
		\includegraphics[width=0.17\linewidth,trim={1.5cm, 0cm, 1.5cm, 0cm}, clip]{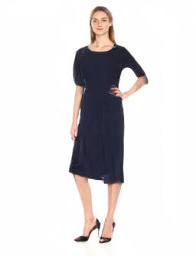}
		\includegraphics[width=0.17\linewidth,trim={1.5cm, 0cm, 1.5cm, 0cm}, clip]{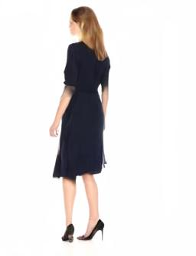}
		\includegraphics[width=0.17\linewidth,trim={0.8cm, 0cm, 2.2cm, 0cm}, clip]{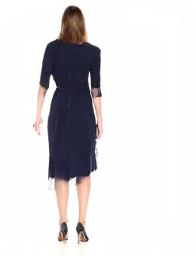}
		\includegraphics[width=0.17\linewidth,trim={0.8cm, 0cm, 2.2cm, 0cm}, clip]{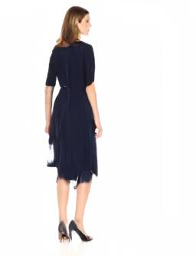}
		\includegraphics[width=0.17\linewidth,trim={1.8cm, 0cm, 1.2cm, 0cm}, clip]{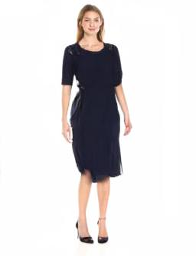}
	}
	\centerline{
        \includegraphics[width=0.17\linewidth,trim={1.5cm, 0cm, 1.5cm, 0cm}, clip]{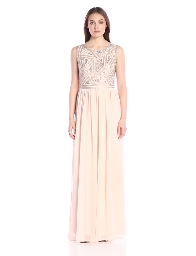} 
		\includegraphics[width=0.17\linewidth,trim={1.5cm, 0cm, 1.5cm, 0cm}, clip]{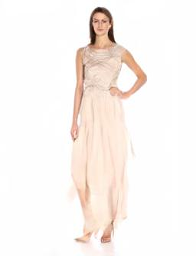}
		\includegraphics[width=0.17\linewidth,trim={1.5cm, 0cm, 1.5cm, 0cm}, clip]{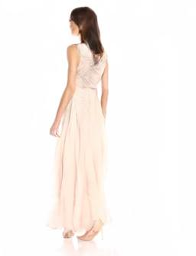}
		\includegraphics[width=0.17\linewidth,trim={0.8cm, 0cm, 2.2cm, 0cm}, clip]{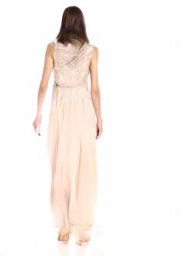}
		\includegraphics[width=0.17\linewidth,trim={0.8cm, 0cm, 2.2cm, 0cm}, clip]{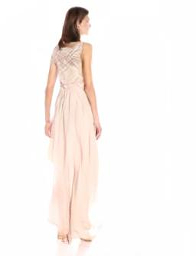}
		\includegraphics[width=0.17\linewidth,trim={1.8cm, 0cm, 1.2cm, 0cm}, clip]{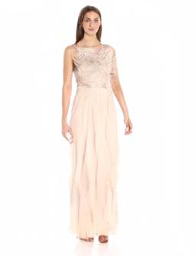}
	}
\vspace{-0.1in}
	\caption{\textbf{Qualitative results} on the Fashion Video dataset \cite{zablotskaia2019dwnet}. Video frames are synthesized from the source frame using the poses in the driving video. See {\bf Sup.~Mat.}~for video and more examples.
	}
	\label{fig:fashion_video_dataset}
\end{figure}

\qheading{Datasets:} 
\modelname is evaluated on two publicly available datasets, namely the DeepFashion In-shop Clothes Retrieval Benchmark \cite{liuLQWTcvpr16DeepFashion} and the Fashion Video datatset \cite{zablotskaia2019dwnet}. 
The DeepFashion data are used for qualitative and quantitative comparisons and Fashion Video data for motion transfer examples, following Sarkar et al.~\cite{Sarkar2020}.  
The DeepFashion dataset \cite{liuLQWTcvpr16DeepFashion} consists of 52712 high-resolution model images in fashion poses. 
The data are split into training and testing sets as in previous work \cite{Ren_2020_CVPR, zhu2019progressive}.
For training, we use 25341 images from the training set, in which body keypoints from nose to knee are at least visible.
Further, 100 randomly selected images from the training set are held out as a validation set for model selection.
The qualitative and quantitative evaluations are performed on the same 8570 image pairs as used by Ren et al.~\cite{Ren_2020_CVPR}.
The Fashion Video dataset \cite{zablotskaia2019dwnet} consists of fashion pose video sequences of women in various clothing, captured with a static video camera.
The dataset is split into 500 training and 100 test videos as done by Sarkar et al.~\cite{Sarkar2020}, with each video containing roughly 350 frames. 
Please note that \modelname uses no paired images for training.

\begin{table}[t]
	\centering
		\begin{tabular}{lccc}
            \toprule
			DeepFashion   & Unpaired & FID($\downarrow$) & LPIPS($\downarrow$) \\ 
			\midrule
			Def-GAN \cite{siarohin2018deformable}       & {\color{red}\xmark}  & 18.5  & 0.233 \\
			Pose-Attn \cite{zhu2019progressive}         &  {\color{red}\xmark} & 20.7  & 0.253 \\
			Intr-Flow \cite{li2019dense}                & {\color{red}\xmark}  & 16.3  & 0.213 \\
			CoCosNet* \cite{zhang2020cross}  & {\color{red}\xmark}   & 14.4 &     -  \\ 
			ADGAN ** \cite{men2020controllable}  & {\color{red}\xmark} & 22.7 &  0.183 \\ 
			Ren et al.~** \cite{Ren_2020_CVPR}             & {\color{red}\xmark}  & 6.4  & 0.143 \\ 
			\midrule
			VUNet ** \cite{esser2018variational}          & {\color{green}\cmark}  & 34.7      & 0.212   \\
			DPIG ** \cite{ma2018disentangled}             & {\color{green}\cmark}  & 48.2       & 0.284   \\
			PGSPT ** \cite{song2019unsupervised} & {\color{green}\cmark} & 29.9 & 0.238 \\
			\textbf{SPICE (Ours)}         & {\color{green}\cmark}         & \textbf{7.8} & \textbf{0.164} \\ 
			\bottomrule
		\end{tabular}
		\vspace{-0.1in}
	\caption{Quantitative comparison of our method with other state-of-the-art methods. The * denotes that the method reports results for a different train/test split.
	The ** denotes that the metrics were recalculated using publicly available code and following the protocol described in {\bf Sup.~Mat.}; note that recalculation of the metrics results in different numbers from those reported in \cite{Ren_2020_CVPR}.}
\vspace{-0.2in}
	\label{table:quantitative}
\end{table}

\qheading{Training details:}
We use residual blocks as basic building blocks for $\generator$.
For more details of the architecture we refer the reader to Ren et al.~\cite{Ren_2020_CVPR}.
We train \modelname with an image resolution of $ 256 \times 256 $ for both datasets.
We use spectral normalisation for both the generator and discriminator.
The learning rate is $8\mathrm{e}{-4}$ for $\generator$ and $1.6\mathrm{e}{-3}$ for the discriminator following similar GAN training strategy of Heusel et al \cite{heusel2017gans}.
We use 8 NVIDIA V100 GPUs to train \modelname, where each GPU has a batch size of 8.
We set the weights for different losses as follows: $\alpha_{cycle} = 1.0, \alpha_{flow} = 1.0, \alpha_{adv} = 1.0, \alpha_{\posecoeff} = 0.01, \alpha_{\shapecoeff} = 0.01, \alpha_{app} = 1.0, \lambda_{a_1} = 0.01, \lambda_{a_2} = 10.0, \lambda_{percep} = 0.5, \lambda_{style} = 500.0, \lambda_{pix} = 5.0$.
First, we train the flow-field estimator.
Differing from Ren et al.~\cite{Ren_2020_CVPR}, we use $\renderingSource$ and $\renderingTarget$ together with keypoints due to the unavailability of $\imageTarget$ during training.
$\renderingSource$ and $\renderingTarget$ are used as the replacement for $\imageSource$ and $\imageTarget$ respectively in their flow estimator module.
We also finetune the 3D regressor $f_{3D}$ on our training splits of the DeepFashion dataset~\cite{liuLQWTcvpr16DeepFashion} following a similar approach of Kolotouros et al.~\cite{Kolotouros_2019_ICCV}. 
During the finetuning of $f_{3D}$, we use a similar representation proposed by Zhou et al.~\cite{zhou2019continuity} for representing 3D rotations.
Finally, we train the whole \modelname model end-to-end keeping the 3D regressor weights fixed.
During a training iteration we use ROIAlign \cite{he2017mask} to extract the desired regions from $\imageSource$ and $\imageSynth$.
We trained our models for 5 days ($\sim$400 epochs). Inference for a single image takes 74 ms using a single NVIDIA V100 GPU.

\qheading{Evaluation metrics:} 
Following Ren et al.~\cite{Ren_2020_CVPR}, we use Learned Perceptual Image Patch Similarity (LPIPS) \cite{zhang2018perceptual} and FID \cite{heusel2017gans} scores to evaluate our experimental results. 
LPIPS quantifies the perceptual distance between the generated image and the ground-truth image. 
The FID score is defined as the Wasserstein-2 distance between the distributions of real and generated images.
We utilize the LPIPS score to evaluate the reconstruction error of \modelname, and the FID score to quantify the realism of the generated images.
Image compression (e.g.~JPEG) applied on reference or generated images significantly affects the FID scores. See {\bf Sup.~Mat.}~for more details or \cite{parmar2021cleanfid} for a similar analysis.
We also evaluated other metrics like contextual similarity \cite{men2020controllable} and object keypoint similarity ~\cite{OKS} and provide the results in {\bf Sup.~Mat.}
Details about the protocol for computing FID and LPIPS can also be found in {\bf Sup.~Mat.}

\qheading{Quantitative evaluation:} \cref{table:quantitative} quantitatively compares our method and other state-of-the-art approaches on the DeepFashion dataset \cite{liuLQWTcvpr16DeepFashion}. 
We compare with Def-GAN \cite{siarohin2018deformable}, Pose-Attn \cite{zhu2019progressive}, Intr-Flow \cite{li2019dense}, CoCosNet~\cite{zhang2020cross}, ADGAN~\cite{men2020controllable}, Ren et al.~\cite{Ren_2020_CVPR}, DPIG \cite{ma2018disentangled}, VUNet \cite{esser2018variational} and PGSPT~\cite{song2019unsupervised}. 
Note that Def-GAN, Pose-Attn, Intr-Flow, CoCosNet, ADGAN and Ren et al.~are supervised methods, requiring
the ground-truth image of a person in the target pose and clothing during training. 
In contrast, our method is unsupervised and comparable with the bottom half of the table (i.e.~\cite{ma2018disentangled, esser2018variational}). 
We have used the publicly available code provided by Ren et al.~\cite{Ren_2020_CVPR}, ADGAN~\cite{men2020controllable}, VUNet \cite{esser2018variational}, DPIG~\cite{ma2018disentangled} and PGSPT~\cite{song2019unsupervised} to regenerate the images on our test split and recompute the metrics.
\modelname achieves state-of-the-art results among unpaired methods and competitive results when compared with supervised methods.

\begin{figure}[t]
    \hspace{0.01\columnwidth} Source \hspace{0.005\columnwidth} Target \hspace{0.005\columnwidth} SPICE \hspace{0.01\columnwidth} DPIG \hspace{0.01\columnwidth} VUNet \hspace{0.01\columnwidth} PGSPT
    \centerline{
        \includegraphics[width=0.48\linewidth]{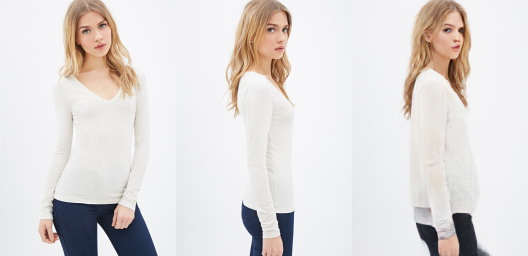} 
        \includegraphics[width=0.16\linewidth]{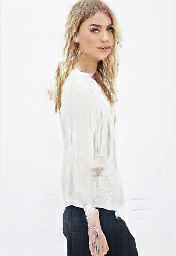} 
        \includegraphics[width=0.16\linewidth]{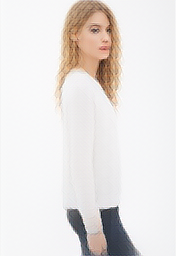} 
        \includegraphics[width=0.16\linewidth]{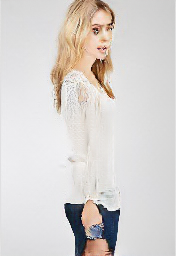} 
        
    }
    \centerline{
        \includegraphics[width=0.48\linewidth]{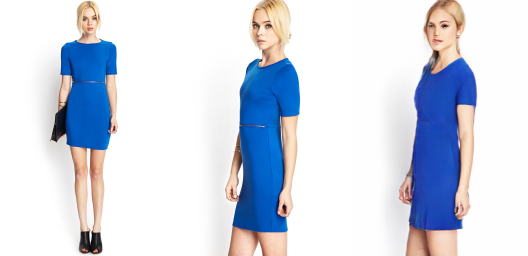} 
        \includegraphics[width=0.16\linewidth]{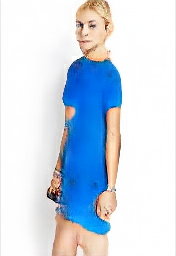} 
        \includegraphics[width=0.16\linewidth]{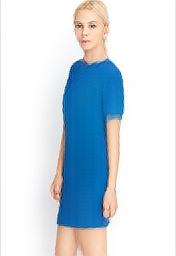} 
        \includegraphics[width=0.16\linewidth]{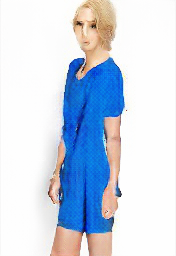} 
        
    }
    \centerline{
        \includegraphics[width=0.48\linewidth]{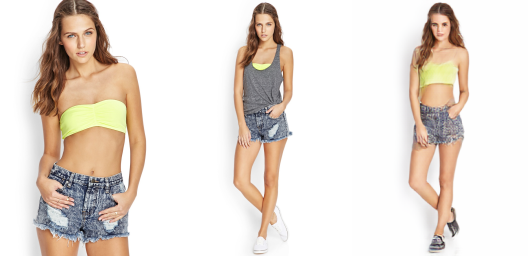} 
        \includegraphics[width=0.16\linewidth]{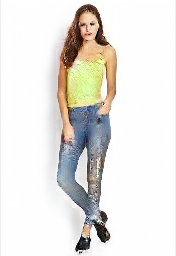} 
        \includegraphics[width=0.16\linewidth]{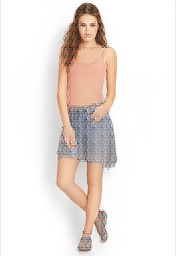} 
        \includegraphics[width=0.16\linewidth]{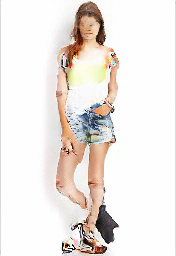} 
        
    }
\vspace{-0.2in}
	\caption{\textbf{Qualitative comparison:} More  results can be found in the {\bf Sup.~Mat.} %
	}
\vspace{-0.1in}
	\label{fig:qualitative_comparison}
\end{figure}

\qheading{Qualitative evaluation:}
Figure \ref{fig:teaser} shows results on the DeepFashion test split.
\modelname does a good job of preserving the target appearance and pose despite the large pose change.
Figure \ref{fig:qualitative_comparison} provides a qualitative comparison with other un/self-supervised methods on the DeepFashion test split.
\modelname generates more realistic and high quality images while preserving pose and appearance compared with DPIG~\cite{ma2018disentangled}, VUNet~\cite{esser2018variational} and PGSPT~\cite{song2019unsupervised}.
See {\bf Sup.~Mat. Video} for more visual results on the DeepFashion test split.

\qheading{Motion transfer:}
If you can generate one pose, you can generate a sequence of poses.
Consequently, we show video generation on the test split of the Fashion Video dataset in Fig.~\ref{fig:fashion_video_dataset}.
We randomly select one video from the test split to act as a driving video that provides the target pose.
We take the first frame of the other videos from the test split as the source image and generate the whole sequence from these.
Please note that we did not train \modelname to generate videos; i.e.~there is no video supervision or temporal consistency.
See {\bf Sup.~Mat. Video}~for examples of generated videos.

\begin{figure}[t]
    \centerline{
        \includegraphics[width=1.0\linewidth]{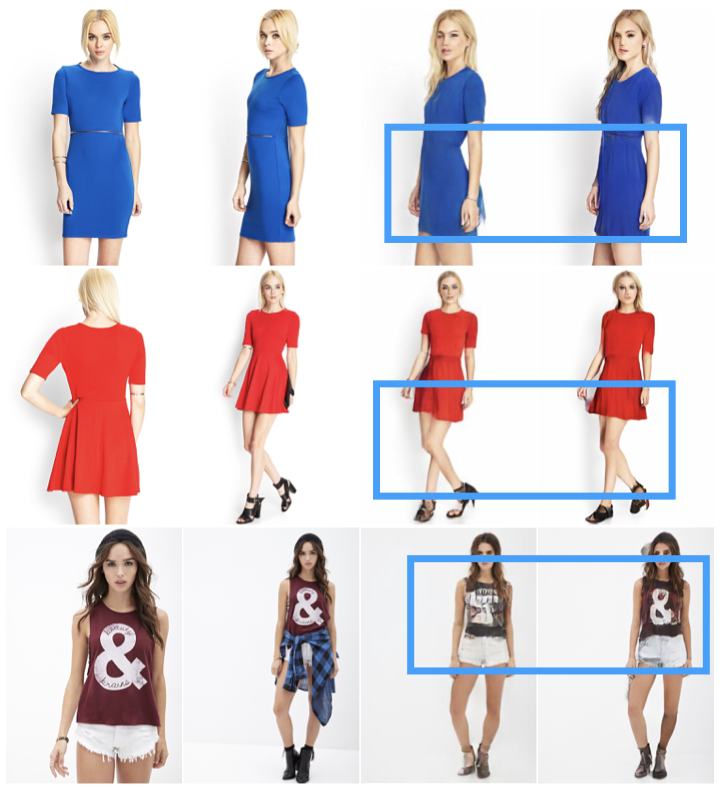}
}
\vspace{-0.1in}
    \caption{\textbf{Loss specific artifacts:} 
    Each row shows artifacts when training without a specific loss.
    Top: without shape loss.
    Middle: without pose loss.
    Bottom: without appearance loss.
    From left to right: source image, reference image in the target pose, generated without the corresponding loss, and \modelname, respectively.}
\vspace{-0.1in}
	\label{fig:ablation_loss_specific}
\end{figure}

\qheading{Ablation study:}
\cref{table:ablationstudy} summarizes our ablation study, which removes one loss at a time from the model.
The configuration ``SPICE w unconditional $D$" means that we
give the generated image to the discriminator without conditioning on pose by concatenating the renderings.
Our full model better preserves details, pose, and has better overall image quality.
Trained without the pose loss, the generator has less information about the self-occlusions of the body.
Therefore it tends to generate poses that are not possible for a real person, e.g.~growing legs inside another leg, etc.
If we train \modelname excluding the shape loss, the generator has less information about the 3D body shape of the person in the source image, which can lead to inconsistent deformations of shape in the generated images; e.g.~having bigger hips with a very thin waist, etc.
Excluding the appearance loss during training leads to less detailed reconstructions and an overall reduced clothing consistency.
\cref{fig:ablation_loss_specific} and the {\bf Sup.~Mat. Video} illustrate such loss-specific artifacts.

\begin{table}[]
\centering
    \begin{tabular}{lcc}
        \toprule
        Configuration                                                                     & FID($\downarrow$)  & LPIPS($\downarrow$) \\ 
        \midrule
        \begin{tabular}[c]{@{}l@{}}SPICE w/o shape loss\end{tabular}       &  8.7  & 0.166 \\
        \begin{tabular}[c]{@{}l@{}}SPICE  w/o pose loss\end{tabular}       &  8.4  & 0.165 \\ 
        \begin{tabular}[c]{@{}l@{}}SPICE w/o appearance loss\end{tabular}  &  9.9  & 0.164 \\ 
        \begin{tabular}[c]{@{}l@{}}SPICE w unconditional $D$ \end{tabular} & 10.0  & 0.167 \\ 
        SPICE                                                              &  7.8 & 0.164 \\ 
        \bottomrule
    \end{tabular}
    \vspace{-0.1in}
    \caption{Ablation study on DeepFashion test set \cite{liuLQWTcvpr16DeepFashion}.
    \vspace{-0.1in}}
    \label{table:ablationstudy}
\end{table}

\begin{figure}
    \centerline{
	    \includegraphics[width=0.5\linewidth]{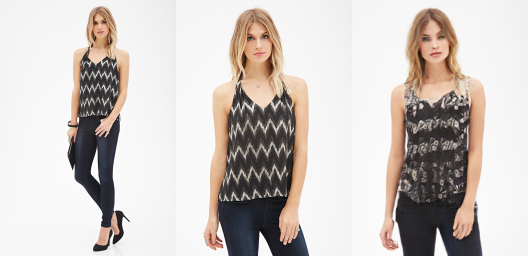}
	    \includegraphics[width=0.5\linewidth]{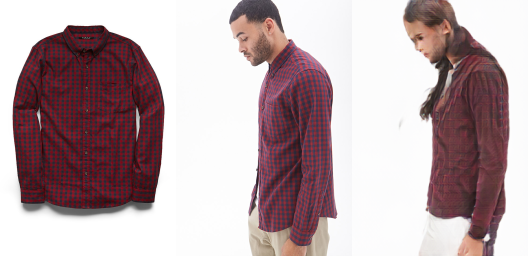}
	}
    \centerline{
        \includegraphics[width=0.5\linewidth]{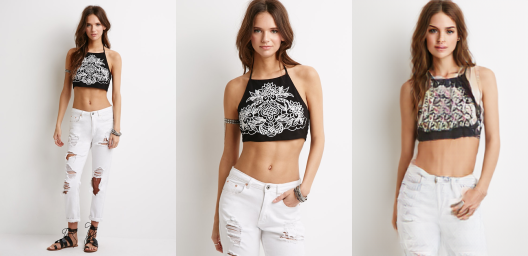}
        \includegraphics[width=0.5\linewidth]{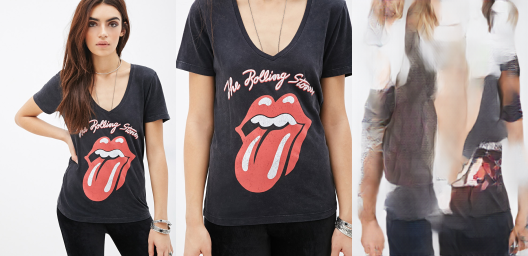}
    }
    \vspace{-0.1in}
    \caption{\textbf{Limitations:} \modelname has difficulty super-resolving fine details when zooming, dealing with extreme closeups, and generating humans from clothing images without humans.}
    \vspace{-0.1in}
	\label{fig:limitations}
\end{figure}

\qheading{Discussions and limitations:}
While the DeepFashion dataset \cite{liuLQWTcvpr16DeepFashion} provides paired data, these pairs do not always have the same outfit, as can be seen in the bottom row
of Figure~\ref{fig:qualitative_comparison}.
We manually checked 500 random sampled pairs of the training data, and found that in 16\% of the pairs one of the images contains additional accessories or new clothes. 
This can be a burden for fully supervised methods. 
Instead, we take the extreme approach of pure self-supervision to see how far this can be pushed. 
For extreme pose/view changes, the solution is highly ambiguous: there is no way to know the front of an outfit from the back or vice versa.
Although \modelname generates a plausible solution, the result might not match the real invisible details.
A practical use case would limit the range of pose variation between source and target.
\modelname requires the target image of a person where much of the body is in view  (Fig.~\ref{fig:limitations}).
Our model has difficulties preserving fine patterns when the camera zooms in Fig.~\ref{fig:limitations}.
Zoom requires super-resolution which is a research topic in itself.

%% file: conclusion.tex
We have presented \modelname, a novel approach to repose clothed humans from a single image. 
\modelname is trained in a self-supervised fashion without paired training data by exploiting cyclic consistency.
Our key insight is to use 3D body information during training in different ways to constrain the image generation.
First, \modelname leverages a parametric 3D body model and a 3D body regressor to constrain body shape and pose.
Second, \modelname uses the 3D body mesh to coherently segment source and generated images to enforce an occlusion-aware appearance feature consistency. 
Third, \modelname conditions a discriminator on colored mesh renderings to increase the quality of the generated images. 
Once trained, \modelname takes a single image and a target pose specified by 2D keypoints, and generates an image of the same person in the target pose.
\modelname generates images that are significantly better than previous unsupervised methods, and that are similar in quality to the state-of-the-art supervised method.
Additionally, \modelname can readily generate videos, although it is not trained for this task.

Adding 3D constraints to the reposing problem enables a number of applications that go beyond the scope of this paper and belong to future work. Although we used our shape and appearance losses to keep those traits constant, they could as well be used to control the output model appearance (e.g.~changing the pattern of a T-shirt) or shape (e.g.~changing the body proportions of the model).

{\footnotesize
\qheading{Disclosure.}
While MJB is also an employee of the Max Planck Institute for Intelligent Systems (MPI-IS), this work was performed solely at Amazon where he is a part time employee. At MPI-IS he has received research gift funds from Intel, Nvidia, Adobe, Facebook, and Amazon. He has financial interests in Amazon, Datagen Technologies, and Meshcapade GmbH.
While TB is a part-time employee of Amazon, his research was performed solely at, and funded by, MPI. }

%% file: supmat.tex
\appendix
\section{FID and LPIPS evaluation}

We pad all generated images to size 256x256 with white border.
Reference images are obtained by resizing the original images from DeepFashion dataset to height of 256 and then padding them to size 256x256 with white border.
We use PyTorch implementations of FID \cite{Seitzer2020FID} and LPIPS \cite{zhangLPIPSCode} with AlexNet as feature extractor. 
The FID is calculated using training images as reference distribution, where generated images are generated using the test split from Ren et al.~\cite{Ren_2020_CVPR} and Zhu et al.~\cite{zhu2019progressive}.

\section{FID for different JPEG quality levels}

Common practice to calculate metrics on generated images is to save the images on disk in JPEG format as an intermediate step. 
We noticed that this affects the FID calculation significantly, as shown in Table~\ref{table:fid_jpeg}. 
The FID increases when it is calculated on image distributions with different levels of JPEG quality and decreases if it is calculated on higher levels of JPEG compression applied to both distributions.

\begin{table}[th]
	\centering
		\begin{tabular}{lcccc}
		\toprule
        \diagbox{GEN}{REF} &  80 &    90 &  95 & \UNCOMPRESSED\\\hline
                     80    &  6.9 &  7.3 &  8.1 &         12.1 \\\hline
                     90    &  7.5 &  7.1 &  7.4 &         10.6 \\\hline
                     95    &  8.7 &  7.8 &  7.4 &          9.6 \\\hline
           \UNCOMPRESSED   & 12.4 & 10.4 &  9.1 &     \bf{7.8} \\
        \bottomrule
		\end{tabular}
		\vspace{-0.1in}
	\caption{FID as a function of the JPEG quality level for generated (GEN) and reference images (REF).}
	\label{table:fid_jpeg}
\end{table}

\section{Additional metrics}

\begin{table}[th]
	\centering
		\begin{tabular}{lccc}
            \toprule
			DeepFashion   & CX-GS($\uparrow$)  & CX-GT($\uparrow$) & OKS($\uparrow$) \\ 
			\midrule
			VU-Net \cite{esser2018variational}          & 0.182 & 0.245 & 0.93 \\
			DPIG \cite{ma2018disentangled}             & 0.164 & 0.197 & 0.86 \\
            PGSPT \cite{song2019unsupervised} &  0.169 & 0.222 & 0.90\\
			\textbf{SPICE (Ours)}                       & \bf{0.236} & \bf{0.311} & \bf{0.94} \\ 
			\bottomrule
		\end{tabular}
		\vspace{-0.1in}
	\caption{Additional quantitative comparison of our method with other unpaired state-of-the-art methods.}
	\label{table:quantitative_additional}
\end{table}

To assess the similarity between the source and generated image and target and generated image we calculate CX scores \cite{men2020controllable}. This score measures the cosine similarity between deep features extracted using VGG19 model between two not aligned images.
We used the original implementation of \cite{CXSCORE}.

Another important metric is the distance between the target pose and pose on the generated image, which can be evaluated using object keypoint similarity (OKS) \cite{OKS}. We used OpenPose \cite{8765346} to extract the keypoints from target and from generated images.

Additional metrics are shown in Table~\ref{table:quantitative_additional}. \modelname outperforms other unsupervised methods on both CX scores and OKS.